%% file: main.tex
\PassOptionsToPackage{numbers,sort&compress}{natbib}
\documentclass{article}

\usepackage[preprint]{neurips_2026}

\usepackage[utf8]{inputenc}
\usepackage[T1]{fontenc}
\usepackage[colorlinks, bookmarks]{hyperref}
\usepackage{url}
\usepackage{booktabs}
\usepackage{amsfonts}
\usepackage{nicefrac}
\usepackage{microtype}
\usepackage{xcolor}
\usepackage{algorithm}
\usepackage{algorithmic}
\usepackage{amsmath}
\usepackage{multirow}
\usepackage{graphicx}
\usepackage{subcaption}
\usepackage{tikz}

\usepackage{wrapfig}
\usepackage{enumitem}

\usetikzlibrary{positioning,arrows.meta,fit,calc}
\title{
SparseForge: Efficient Semi-Structured LLM Sparsification via Annealing of Hessian-Guided Soft-Mask
}

\author{
  \textbf{Hanzuo Liu}$^{\dag\ddag}$ \quad
  \textbf{Chaofan Lin}$^{\dag}$ \quad
  \textbf{Weixuan Sun}$^{\ddag}$ \quad
  \textbf{Yulong Wang}$^{\ddag}$\\
  \textbf{Rayying}$^{\ddag}$ \quad
  \textbf{Key}$^{\ddag}$ \quad
  \textbf{Mingyu Gao}$^{\dag}$ \\
$^{\dag}$\text{Tsinghua University} \quad 
$^{\ddag}$\text{Tencent} \\
\texttt{\{lhz24@mails., lcf24@mails., gaomy@\}tsinghua.edu.cn} \\
\url{https://github.com/tsinghua-ideal/SparseForge}
}

\begin{document}

\maketitle
\input{sections/abstract}
\input{sections/introduction}
\input{sections/related_work}

% New section added for logical clarity
\input{sections/motivation}

\input{sections/methodology}
\input{sections/experiments}
\input{sections/conclusions}
\newpage
\bibliographystyle{plainnat}
\bibliography{references}

\input{sections/appendix}

\end{document}

%% file: sections/abstract.tex
\begin{abstract}
Semi-structured sparsity provides a practical path to accelerate large language models (LLMs) with native hardware support, but post-training semi-structured pruning often suffers from substantial quality degradation due to strong structural coupling. Existing methods rely on large-scale sparse retraining to recover accuracy, resulting in high computational cost.

We propose SparseForge, a post-training framework that improves recovery efficiency by directly optimizing the sparsity mask rather than scaling up retraining tokens.
SparseForge combines Hessian-aware importance estimation with progressive annealing of soft masks into hardware-executable structured sparsity, enabling stable and efficient sparse recovery.
On LLaMA-2-7B under 2:4 sparsity, SparseForge achieves 57.27\% average zero-shot accuracy with only $\textbf{5B}$ retraining tokens, surpassing the dense model's 56.43\% accuracy and approaching the 57.52\% result of a state-of-the-art method using $\textbf{40B}$ tokens. 
Such improvements on the accuracy-efficiency trade-off from SparseForge are shown to be consistent across model families.
\end{abstract}

%% file: sections/introduction.tex
\section{Introduction}
% 引子 e.g. 剪枝的部署, 但是要突出 soft mask(让读者直接知道文章的独特性)

Semi-structured sparsity stands out as one of the few sparsity paradigms that deliver both meaningful compression and tangible hardware acceleration for large language models (LLMs). In particular, patterns such as 2:4 sparsity are natively supported by modern NVIDIA Sparse Tensor Core pipelines, making them attractive for practical deployment~\citep{nvidia2020ampere}. However, in post-training LLM sparsification, achieving high accuracy under such structured constraints typically requires substantial retraining, leading to a nontrivial trade-off between performance and training cost.

Existing post-training pruning methods often rely on computing importance scores for weights followed by hard projection to sparse structures~\citep{frantar2023sparsegpt,sun2023wanda,huang2025ast,huang2025cast}. While effective in simpler settings, this paradigm becomes brittle under semi-structured constraints, where pruning decisions are no longer independent. Instead, weights compete within each constrained group, and premature hard decisions can remove weights that are individually important but suboptimal under imperfect scoring. As a result, these approaches either suffer from noticeable accuracy degradation or require large amounts of retraining to recover performance.

\begin{figure*}[t]
    \centering
    \begin{subfigure}[b]{0.67\textwidth}
        \centering
        \includegraphics[width=\textwidth]{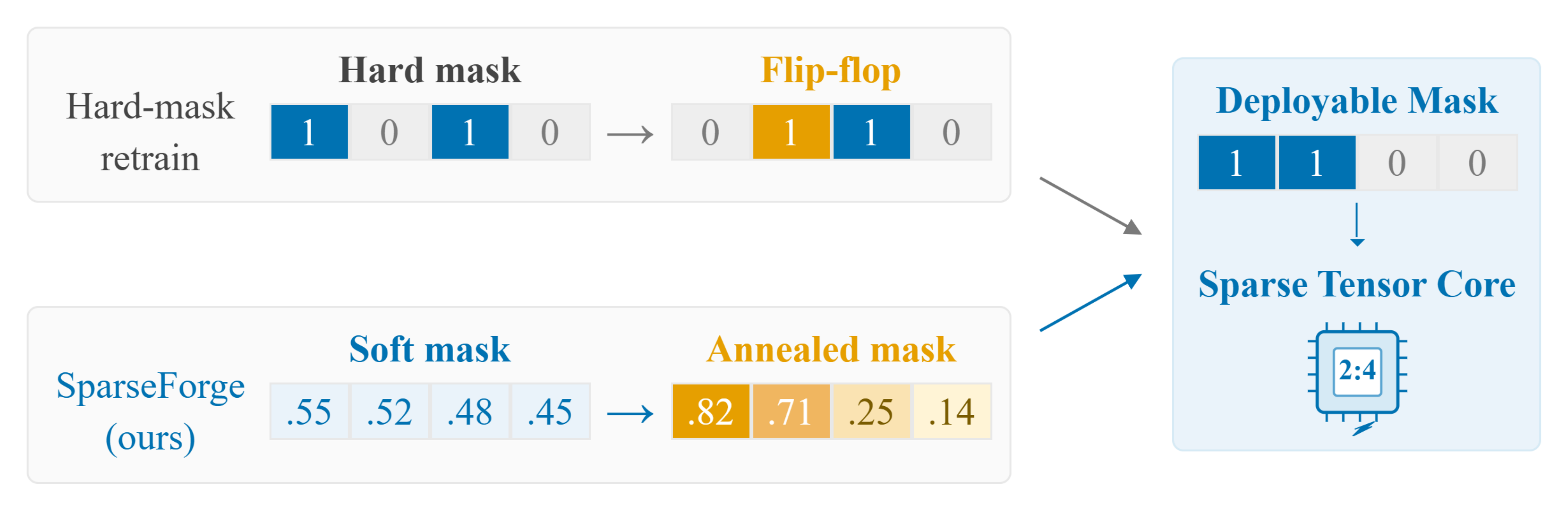}
        % \caption{SparseForge: soft masks tempered by Hessian to deployable 2:4.}  
        \label{fig:teaser_a}
    \end{subfigure}
    \hfill
    \begin{subfigure}[b]{0.31\textwidth}
        \centering
        \includegraphics[width=\textwidth]{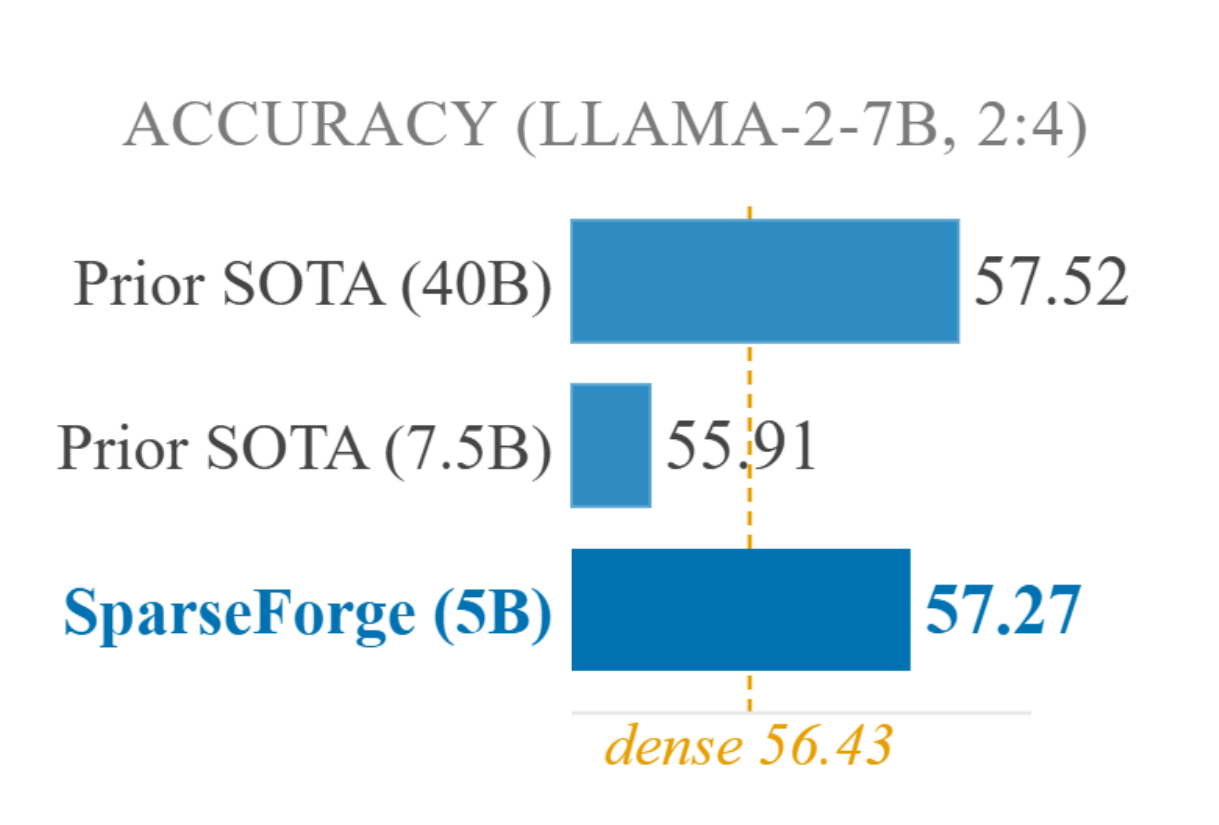}
        % \caption{Beats dense LLaMA-2-7B with $8\times$ fewer tokens.}
        \label{fig:teaser_b}
    \end{subfigure}
    \caption{
    \textbf{(a)} Compared with hard-mask retraining, SparseForge explicitly optimizes a soft mask and progressively anneals it into a deployable binary 2:4 mask.
    \textbf{(b)} On LLaMA-2-7B under 2:4 sparsity, SparseForge achieves 57.27\% average zero-shot accuracy with 5B retraining tokens, approaching the 57.52\% result of the stronger 40B-token prior SOTA variant~\citep{huang2025cast} while using about $8\times$ fewer training tokens.
    }
    \label{fig:teaser}
\end{figure*}

We argue that this challenge is fundamentally rooted in \emph{how sparsity masks are learned}. Rather than committing to a discrete mask prematurely, semi-structured sparsification should treat the mask as a continuous and optimizable variable, allowing the model to explore structured alternatives before making irreversible pruning decisions. More importantly, we argue that improving mask optimization quality can reduce reliance on large-scale retraining, offering a more efficient path to sparse recovery. Adopting a continuous mask formulation, however, introduces a new challenge: at inference time the mask must be an exact, hardware-executable binary pattern, creating a gap between the learned soft mask and its deployment-time hard counterpart. In practice, naive soft-mask approaches often lead to performance degradation when the mask is binarized, due to insufficient separation between mask values.

Based on this perspective, we propose \textbf{SparseForge} (Figure~\ref{fig:teaser}), a post-training framework for semi-structured sparsification that leverages \emph{Hessian-aware importance} guideline to derive a \emph{soft mask}, and progressively hardens it by gradually shaping from continuous exploratory values into deployable binary sparse structures. 
SparseForge improves mask optimization quality via three key techniques: (1) continuous soft-mask optimization for stable structured exploration; (2) Hessian-aware importance estimation to better capture deletion sensitivity under grouped competition; and (3) a progressive annealing mechanism that gradually aligns soft masks with hardware-executable binary patterns. With improved mask quality, SparseForge largely reduces retraining tokens and hence achieves efficient sparse recovery. The main contributions of this work are as follows:
\begin{itemize}[leftmargin=2em]
\item We identify that under semi-structured constraints, sparse recovery is bottlenecked more by mask quality than by retraining scale. 
This insight motivates us to prioritize optimizing the mask itself and to keep it soft during retraining, which leads to the proposed \emph{mask annealing} pipeline that efficiently navigates the soft-to-hard transition.
% : early discrete decisions are brittle under grouped competition, magnitude-only ranking is insufficient to resolve local survival choices, and soft masks that are not progressively tempered incur a large soft-to-hard gap at deployment.

\item We present SparseForge, a post-training framework for semi-structured sparsification that explicitly optimizes the mask itself. SparseForge features a Hessian-aware mask annealing algorithm, which smoothly trains the soft mask towards a hardware-executable binary pattern.

\item We show that SparseForge attains accuracy on par with both the dense model and state-of-the-art sparse methods with \emph{remarkably fewer retraining tokens} (e.g., 5B vs. 40B on LLaMA-2-7B).
% is effective across model families, yielding strong sparse recovery on diverse architectures and scales. 
% On LLaMA-2-7B under 2:4 sparsity, it achieves a favorable accuracy--efficiency trade-off, surpassing the dense baseline and approaching prior state-of-the-art recovery with substantially fewer retraining tokens. 
Extensive ablations further verify the contribution of its key components.
\end{itemize}

%% file: sections/related_work.tex
\section{Related Work}
\label{sec:related}

\paragraph{Hardware-aligned sparsity patterns.}
Structured sparsity has long been pursued as a practical route to efficient neural network acceleration, since regular sparse patterns are far easier for hardware and kernels to exploit than irregular ones~\citep{wen2016learning,li2017pruning,liu2017network,gale2019state,mishra2021accelerating}. Early work explored filter-, channel-, and block-level pruning for real execution benefits. More recently, semi-structured patterns such as 2:4 sparsity have drawn significant attention due to native GPU support~\citep{nvidia2020ampere}. While offering clear deployment gains, such patterns impose tightly coupled local constraints---selecting exactly $N$ survivors within each group of $M$ (in practice $N{=}2$, $M{=}4$, i.e., 2:4 sparsity)---which makes their accuracy impact more substantial than flexible, unstructured cases.

\paragraph{Unstructured sparsification for LLMs.}
A large body of recent work studies post-training sparsification for LLMs in the unstructured setting. Building on classical magnitude-based pruning~\citep{han2015magnitude}, representative methods such as SparseGPT~\citep{frantar2023sparsegpt} and Wanda~\citep{sun2023wanda} estimate weight importance via approximate second-order information or activation statistics, removing parameters with little or no retraining. OWL~\citep{yin2024owl} further shows that outlier-weighed, layerwise sparsity allocation is critical for pushing LLMs to high sparsity. BESA~\citep{xu2024besa} introduces a blockwise sparsity allocation strategy, but its goal is still not hardware-aligned semi-structured execution. However, unstructured sparsity does not translate into efficient execution on current accelerators, so it does not resolve the deployment problem targeted by hardware-aligned patterns.
% \mingyu{Are LLM-Pruner and BESA unstructured? From the above description (dependency-aware structural removal, blockwise differentiable), there seem to be some structuring?}

\paragraph{Structured sparsification for LLMs.}
Recent efforts therefore turn to structured or semi-structured sparsification for LLMs. MaskLLM~\citep{fang2024maskllm} cast $N$:$M$ mask selection as a learnable discrete distribution optimized via Gumbel-Softmax sampling on large-scale corpora, while AST~\citep{huang2025ast} followed a dense-to-sparse adaptive sparse training scheme that progressively annealed the weights toward the 2:4 pattern. More recently, CAST~\citep{huang2025cast} relaxed the hard 2:4 constraint into a continuous, differentiable form so that masks could be optimized jointly with weights. These works show that better mask optimization can substantially narrow the gap of hardware-executable sparse models, yet
% ---despite their differing treatments of the $N$:$M$ constraint---
they all rely on training over a large number of tokens for the mask to converge to a high-quality solution. Our work instead targets \emph{recovery efficiency} by improving mask optimization quality itself, attaining strong recovery with considerably fewer retraining tokens.

%% file: sections/motivation.tex
    \section{Motivation}
    \label{sec:motivation}

    \subsection{Rethinking the Bottleneck of Sparse Recovery: Mask Quality over Retraining Scale}
    
    As discussed in \S\ref{sec:related}, previous methods such as CAST~\citep{huang2025cast} preserve quality by scaling retraining tokens, which yields gains but with poor efficiency. Observations from CAST show that even with substantial retraining, accuracy is still governed by the selected sparse pattern, implying that \emph{mask quality, rather than retraining scale alone, is the key bottleneck.} We instead choose to directly improve the quality of mask optimization along three axes: (1) keeping the mask soft long enough to avoid brittle early hard decisions, (2) using Hessian-aware importance to resolve intra-group competition, and (3) progressively annealing the mask into the final hardware-executable $2{:}4$ pattern. 
    
    \subsection{Soft Masks are Necessary under Grouped Semi-structured Constraints}
    
    % single-side figure through wrapfigure
    \begin{wrapfigure}[16]{r}{0.45\textwidth}
    \centering
    \vspace{-20pt}
    \includegraphics[width=0.4\textwidth]{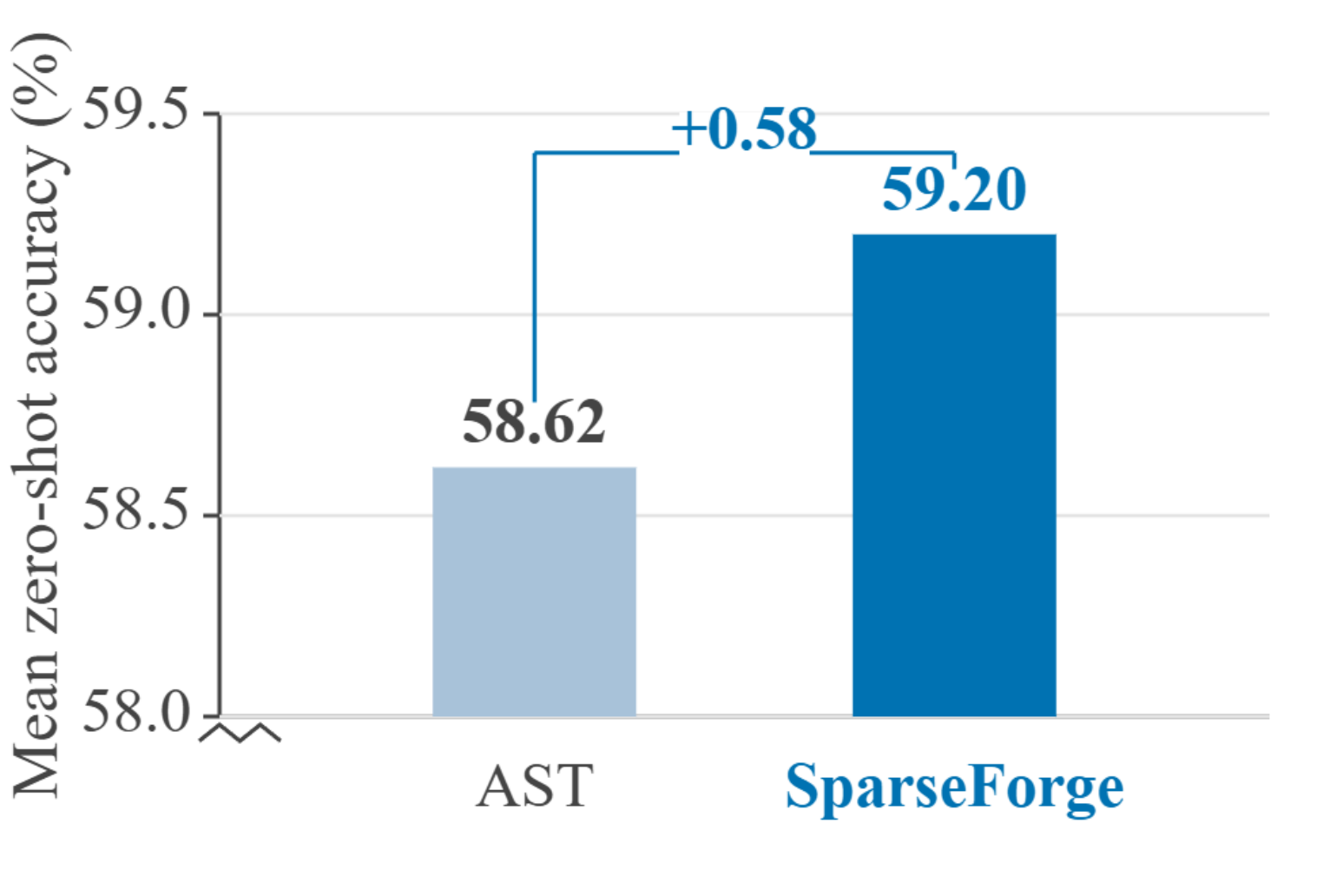}
    \vspace{-12pt}
    \caption{
    Soft masks improve sparse recovery: compared with the hard-mask-style AST baseline, our soft-mask-style SparseForge improves the mean zero-shot accuracy on the 7-task benchmark from 58.62\% to 59.20\%, supporting the need for soft mask optimization.
    }
    \label{fig:ppl}
    \end{wrapfigure}
    
\begin{figure*}[htbp]
        \centering
        % (b) hessian-aware
        \begin{minipage}[t]{0.49\textwidth}
            \centering
            \includegraphics[width=\textwidth]{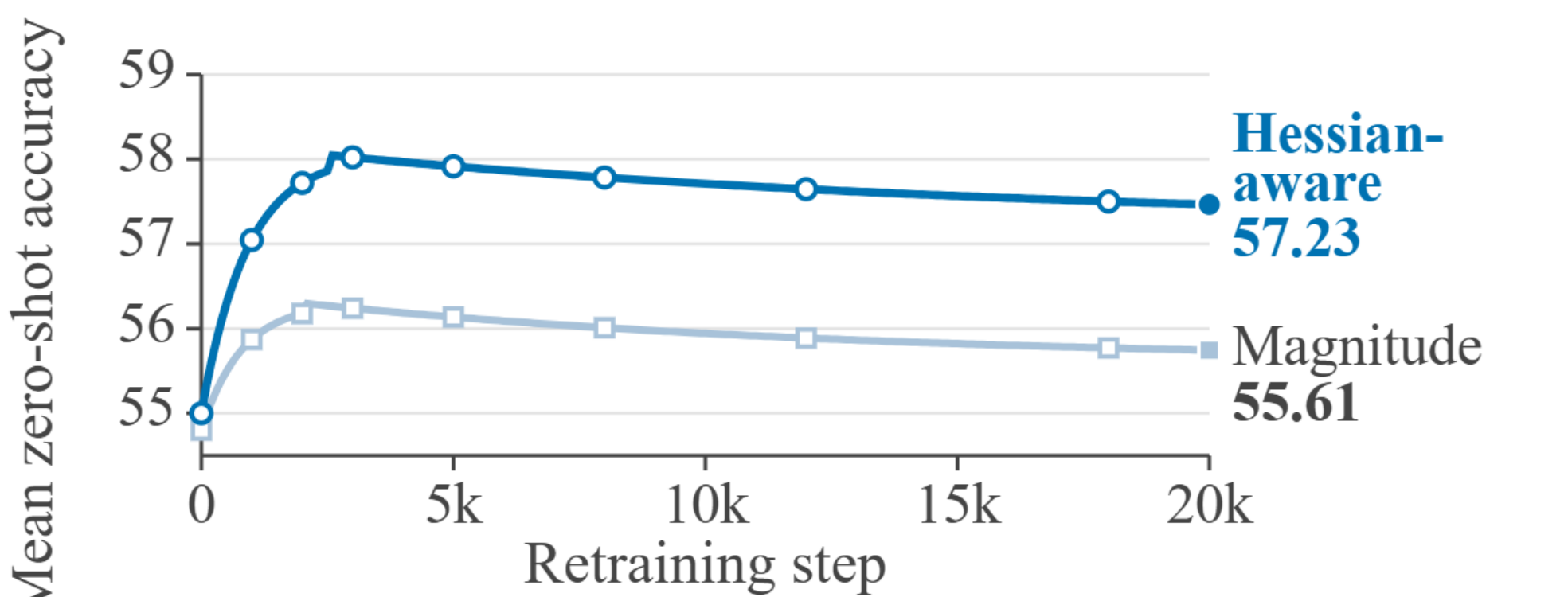}
        \end{minipage}
        \hfill
        % (c) top/bottom stacked
        \begin{minipage}[t]{0.49\textwidth}
            \centering
            \includegraphics[width=\textwidth]{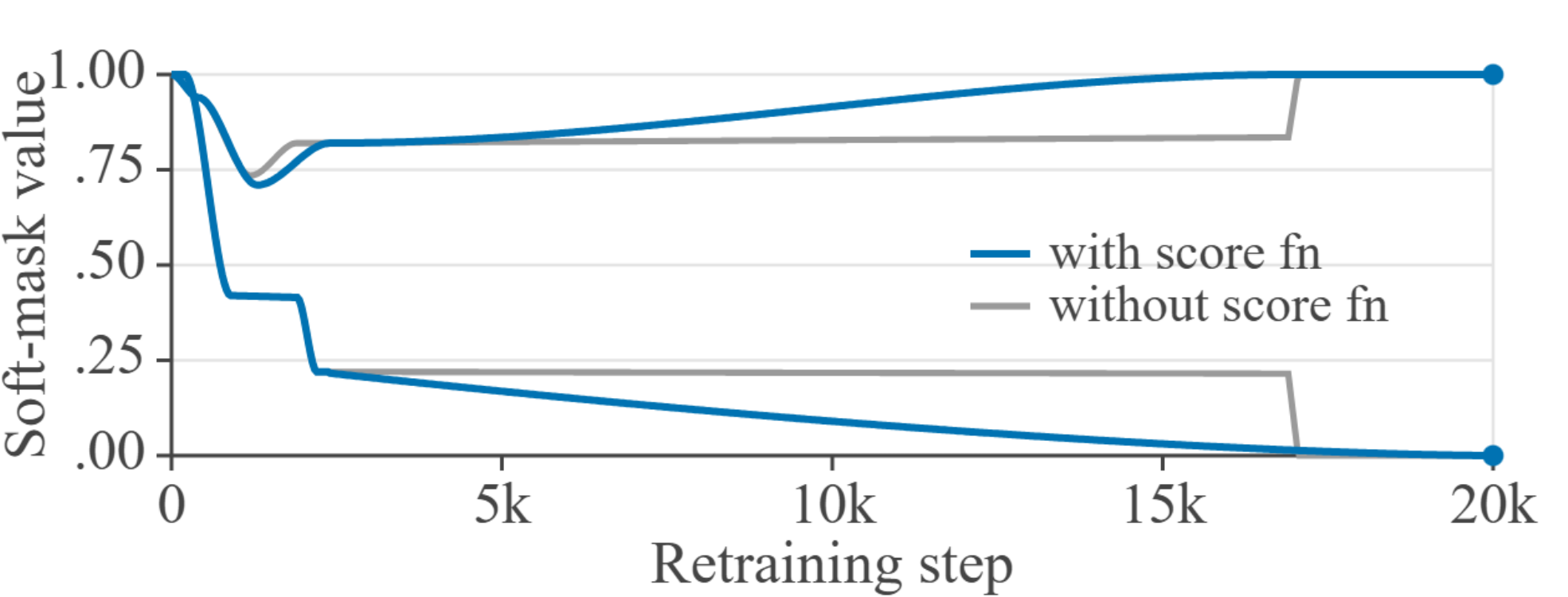}
        \end{minipage}
    
        \caption{
        % \textbf{Empirical motivation of SparseForge.}
        \textbf{(a)} Hessian-aware importance provides a better survival signal: replacing it with magnitude-based scoring drops mean zero-shot accuracy from 57.23\% to 55.61\%.
        \textbf{(b)} Soft masks must also be progressively hardened toward a deployable 2:4 pattern: score structural hardening pushes the top-2 and bottom-2 entries in each group towards 1 and 0, respectively; without such hardening, the mask remains far from binary, leading to a larger soft-to-hard gap at final projection.
        % \mingyu{In (a), why do we have two bars for AST? The text only talks about one, so no one knows what is the other one. Maybe just keeping on bar to compare with SparseForge is enough.}
        % \mingyu{In (b), some text (Hessian-aware) is hidden in the figure. Fix it.}
        % \mingyu{In (c), the legend needs to be better placed, not overlapping with the lines.}
        }
        \label{fig:motivation_summary}
\end{figure*}

    Under semi-structured constraints such as 2:4 sparsity, pruning is no longer an element-wise decision but a grouped selection problem in which weights compete within each constrained block. Committing to a binary mask too early therefore locks the optimizer into a suboptimal sparse subspace before intra-group importance is sufficiently resolved. This effect is also reflected empirically in Figure~\ref{fig:motivation_summary}(a): on the 7-task zero-shot benchmark, SparseForge achieves 59.20\% mean accuracy, outperforming the hard-mask-style AST baseline at 58.62\%~\citep{huang2025ast}. This supports the need to \emph{keep the mask soft} during optimization.
    
    However, a soft mask must eventually be converted into a deployable binary (0/1) form. Combining the two sides, we propose a pipeline that we call \emph{$\textbf{mask annealing}$} in this paper: first, we optimize the mask in a soft, continuous state to allow extensive exploration, analogous to heating metal into a malleable condition (\textbf{Heating}); 
    then, we binarize the mask into the hardware-ready pattern, analogous to cooling it into a fixed shape (\textbf{Quenching}).
    
    \subsection{Insights to Implement Efficient Mask Annealing}
    \label{sec:insights}
    With mask annealing, we appear to achieve the best of both worlds: the exploration flexibility of a soft mask and the deployment readiness of a hard mask. Yet, translating this idea into practice requires several key insights, which we detail below.
    
    \paragraph{(I1) Using second-order Hessian-aware loss signal.} Even with a soft mask, the optimizer still needs to decide which entries should survive within each constrained group. A useful intuition comes from the local loss change incurred by removing a weight:
    \[
    \Delta L = L(w_i=0)-L(w_i)
    \approx \frac{\partial L}{\partial w_i}(-w_i) + \frac{1}{2}\frac{\partial^2 L}{\partial w_i^2} w_i^2
    \]
    Near a converged solution, the first-order gradient term is typically negligible, so removal sensitivity is dominated by the local curvature term. Under grouped competition, magnitude alone is therefore often insufficient to determine which entries should be preserved. Figure~\ref{fig:motivation_summary}(a) directly supports this point: using magnitude-based scoring to replace Hessian-aware importance would drop the mean zero-shot accuracy from 57.23\% to 55.61\%. This is also consistent with classical second-order pruning criteria~\citep{lecun1990obd,hassibi1993obs,singh2020woodfisher} and modern LLM pruning methods such as SparseGPT~\citep{frantar2023sparsegpt}.
    
    \paragraph{(I2) The quenching process needs to be progressive.} Under 2:4 sparsity,  the two surviving entries in each group should approach 1, while the two pruned ones should approach 0. If many mask values remain in the middle of $[0,1]$ in the late stages during training, the final quenching step must abruptly binarize them, causing a large projection error. This motivates us to optimize soft mask towards a near-binary structured solution during the heating stage, rather than relying on a late hard projection alone in the quenching stage. As in Figure~\ref{fig:motivation_summary}(b), with a score function explicitly designed to encourage the mask toward a near-binary state, retraining steadily pushes the entries toward 0/1.

%% file: sections/methodology.tex
\section{Methodology}
% 1.前置training objective. 先说我们的设计和CAST, AST的framework类似. 然后说我们的要点: mask

% \begin{itemize}[leftmargin=2em]
% \item \textbf{Heating stage:} The soft mask modulates weights in the forward pass, allowing flexible exploration. Hessian-guided importance scores construct a structured target that directs mask updates under intra-group competition, providing faithful survival signals.
% \item \textbf{Quenching stage:} A progressive, structure-aware quenching schedule gradually shapes the mask toward a deployable semi-structured pattern, so that the final binary projection is gentle and low-loss.
% \end{itemize}

\begin{figure}[htbp]
    \centering
    \makebox[\textwidth][c]{
        \includegraphics[width=\textwidth]{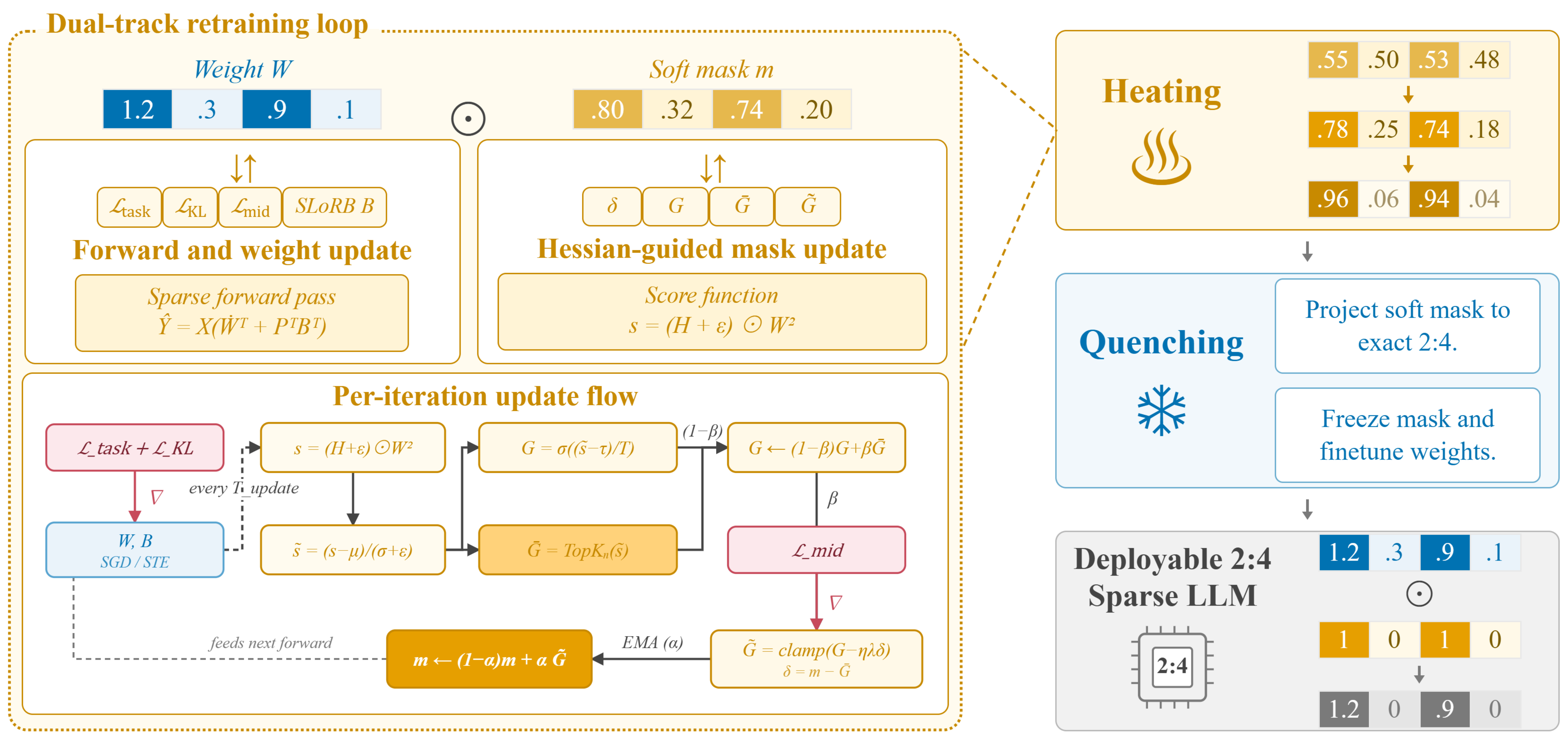}
    }
    \caption{\textbf{Overview of SparseForge.} We first co-optimize the weights and the explicit learnable soft mask in a dual-track retraining loop (\S\ref{sec:retrain_overview}) in the heating stage, where the mask is updated with a Hessian-guided signal (\S\ref{sec:mask_upd}). Then we progressively binarize the mask to a hard, deployable form in the quenching stage (\S\ref{sec:progress}).
    % SparseForge operates in an iterative retraining loop. Given the current soft mask, dense weights are element-wise modulated to produce a sparse forward pass (\textit{left}), optionally with an SLoRB compensation branch. From the current model state, a Hessian-aware importance score $s=(H+\varepsilon)\odot W^2$ is computed and
    % --- ORIGINAL ---
    % projected via $\mathrm{TopK}_2$ to form a structured target, which is then used to update and progressively temper the soft mask (\textit{center}).
    % --- REVISED ---
    % turned into a group-wise soft gate via a temperature-controlled soft threshold, optionally blended with the hard $\mathrm{TopK}_N$ target to inject N:M structure, and used to update and progressively temper the soft mask (\textit{center}).
    % Task loss, KL distillation, and binary-preference regularization jointly supervise this loop. As training proceeds, the mask becomes increasingly near-binary and structurally aligned, so that the final projection and brief mask-frozen finetuning (\textit{right}) yield a deployable 2:4 sparse model with limited additional loss.
    }
    \label{fig:sparseforge_pipeline}
\end{figure}

Building on the insights from \S\ref{sec:motivation}, we design \textbf{SparseForge}, a LLM sparsification framework that efficiently implements the mask annealing pipeline. Figure~\ref{fig:sparseforge_pipeline} provides an overview.
In \S\ref{sec:retrain_overview}, we present the overall retraining loop of the heating stage, describing how the soft mask is optimized alongside model weights. \S\ref{sec:mask_upd} then addresses (I1) by detailing our second-order mask update rule, while \S\ref{sec:progress} tackles (I2) and explains how progressive quenching bridges the soft-to-hard gap.

\subsection{Dual-track Retraining Loop in the Heating Stage}
\label{sec:retrain_overview}

During the heating stage, SparseForge jointly optimizes the model weights and the soft mask in a unified retraining loop. 
The weight update follows the paradigm established by prior sparse-training methods such as AST and CAST~\citep{huang2025ast, huang2025cast}. Given a linear layer
\[
Z = XW^T, \qquad X \in \mathbb{R}^{N\times C},\; W \in \mathbb{R}^{D\times C},\; Z \in \mathbb{R}^{N\times D},
\]
a sparse mask $m \in \{0, 1\}^{D \times C} \cap \mathcal{M}$ is applied in the forward pass by writing $\tilde W = m \odot W$, where $\mathcal{M}$ is the semi-structured feasible set (e.g., 2:4 or block-16). The sparse forward pass, optionally equipped with a sparse low-rank branch (SLoRB)~\citep{huang2025ast}, is
\begin{equation}
\hat Y = X (\tilde W^T + P^TB^T),
\end{equation}
where $P$ is the fixed projection and $B$ is a learnable low-rank parameter used to compensate for capacity lost during pruning. Our retraining objective builds upon prior work and introduces a structure-aware guidance term that facilitates progressive quenching:
\begin{equation}
\mathcal L = \lambda_{\mathrm{task}}\mathcal L_{\mathrm{task}}
+ \lambda_{\mathrm{KL}}\mathcal L_{\mathrm{KL}}
+ \lambda_{\mathrm{mid}}\mathcal L_{\mathrm{mid}},
\end{equation}
% --- ORIGINAL ---
% where $\mathcal L_{\mathrm{task}}$ is the standard retraining task loss, $\mathcal L_{\mathrm{KL}}$ is the distillation loss between the logits of the sparse model and the dense teacher, and $\mathcal L_{\mathrm{mid}}$ is the binary-preference regularization loss (\S\ref{sec:progress}).
% --- REVISED ---
where $\mathcal L_{\mathrm{task}}$ and $\mathcal L_{\mathrm{KL}}$ inherit the standard task and distillation losses from prior work, and $\mathcal L_{\mathrm{mid}}$ extends the objective with a binary‑preference regularizer that we design for the mask (\S\ref{sec:progress}).
% ; its gradient is injected into the mask-update rule (\S\ref{sec:mask_upd}) rather than back-propagated through the forward pass, because $m$ is maintained as a non-differentiable buffer updated by its own rule.

Besides weight optimization, SparseForge performs explicit soft-mask learning. In prior works, masks are typically treated as hard selections derived from one-shot importance scores or updated through discrete re-selection steps, rather than being continuously optimized as learnable variables. 
SparseForge instead separates mask learning from weight adaptation and makes the soft mask the central object of optimization. 

Specifically, we optimize a continuous mask $m \in [0,1]^{D\times C}$ during retraining and only project it to a hardware-executable binary pattern at the end.
Within the same unified loop, the optimization proceeds along two interleaved tracks: the weights $W$ and the low-rank $B$ follow the previous paradigm, updated by the loss at every step, while the soft mask $m$ is updated by its own dedicated learning rule (\S\ref{sec:mask_upd}) only every $T_{\text{update}}$ steps. This decoupled schedule confines sparsification entirely to the mask space, where optimization remains more stable under grouped semi‑structured constraints. We briefly sketch this \emph{dual-track retraining loop} in Algorithm~\ref{algo:training}. The following sections elaborate the details.

\begin{algorithm}[htbp]
\caption{SparseForge's Dual-track Retraining Loop}
\label{algo:training}
\begin{algorithmic}[1]
\REQUIRE Dense weights $W$, semi-structured constraint $\mathcal M$, training data $\mathcal D$, mask update duration $T_{\mathrm{update}}$
% --- ORIGINAL ---
% \STATE Initialize soft mask $m \leftarrow \mathbf 1$, gate $G \leftarrow \mathbf 1$
% --- REVISED ---
\STATE Initialize soft mask $m \leftarrow \mathbf 1$; initialize temperature $T \leftarrow T_0$ and structural-mix factor $\beta \leftarrow 0$
\STATE Initialize optional SLoRB branch $(P,B)$
\FOR{$t=1$ to $T_{\max}$}
    \STATE Sample batch $(X,Y) \sim \mathcal D$
    \STATE Compute sparse forward pass with current soft mask $m$
    % --- ORIGINAL ---
    % \STATE Update $W$ (via STE) and $B$ using $\lambda_{\mathrm{task}}\mathcal L_{\mathrm{task}} + \lambda_{\mathrm{KL}}\mathcal L_{\mathrm{KL}} + \lambda_{\mathrm{mid}}\mathcal L_{\mathrm{mid}}$
    % --- REVISED ---
    \STATE Update $W$ (via STE) and $B$ using $\lambda_{\mathrm{task}}\mathcal L_{\mathrm{task}} + \lambda_{\mathrm{KL}}\mathcal L_{\mathrm{KL}}$ \hfill\COMMENT{$\mathcal L_{\mathrm{mid}}$ enters through \eqref{eq:mid_inject}, not autograd}
    \IF{$t \bmod T_{\mathrm{update}} = 0$}
        \STATE Compute importance score $s=(H+\epsilon)\odot W^2$; standardize to $\tilde s$
        % --- ORIGINAL ---
        % \STATE Build structured target $\bar G$ by projecting $s$ onto $\mathcal M$
        % \STATE Update gate $G$ and blend with $\bar G$ using annealing factor $\beta$
        % \STATE Update soft mask $m \leftarrow (1-\alpha) \cdot m + \alpha \cdot \mathrm{clamp}(G_{\mathrm{final}},0,1)$
        % --- REVISED ---
        \STATE Synthesize soft gate $G = \sigma((\tilde s - \tau)/T)$ with group-wise threshold $\tau$, and hard target $\bar G = \mathrm{TopK}_N(\tilde s)$ on $\mathcal M$
        \STATE Blend in structure: $G \leftarrow (1-\beta)\,G + \beta\,\bar G$
        \STATE Inject mid-penalty: $\tilde G \leftarrow \mathrm{clamp}\!\left(G - \eta_{\mathrm{pen}}\lambda_{\mathrm{mid}}(m - \bar G),\,0,\,1\right)$
        \STATE Update soft mask: $m \leftarrow (1-\alpha)\cdot m + \alpha\cdot \tilde G$
        \STATE Anneal schedules: $T \leftarrow \gamma T$; advance $\beta$, $\lambda_{\mathrm{mid}}$
    \ENDIF
\ENDFOR
% \STATE Interpolate from soft mask to hard mask, finalize structured projection, and finetune briefly
\RETURN Near-hard mask $m$ and the trained model
\end{algorithmic}
\end{algorithm}

\subsection{Hessian-Guided Soft-Mask Update}
\label{sec:mask_upd}

Following (I1) (\S\ref{sec:insights}), we derive the survival signal for mask updates from a \emph{Hessian-aware} criterion, which yields the classic OBD-style importance score~\citep{lecun1990obd}:
\[
s = (H + \epsilon) \odot W^2, \qquad s \in \mathbb{R}^{D\times C},
\]
where $s_{ij}$ measures the Hessian-aware pruning sensitivity of weight $W_{ij}$, $H \in \mathbb{R}^{D\times C}$ denotes an approximate Hessian diagonal reshaped to the weight matrix, and $\epsilon$ is a small stabilizer.

Since explicitly forming the Hessian is prohibitively expensive for LLMs, we estimate its diagonal with a Hutchinson-style stochastic estimator~\citep{hutchinson1990trace}:
\[
\mathbb{E}[(Hv) \odot v] = \mathrm{diag}(H), \qquad v \sim \text{Rademacher}(\pm 1),
\]
and maintain an exponential moving average of $(Hv)\odot v$ in practice.

% --- ORIGINAL ---
% To respect the final semi-structured constraint, we project the Hessian-aware scores onto the feasible set and obtain a structured binary target $t \in \{0,1\}^{D\times C}$. For an $N$:$M$ pattern, let $\mathrm{Groups}(M)$ denote the collection of non-overlapping groups, each containing $M$ mask entries along the structured sparsity dimension. For a group $\mathcal G$, $s_{\mathcal G}$ denotes the restriction of the score matrix $s$ to the entries in $\mathcal G$. We then define
% \[
% t_{\mathcal G} = \mathrm{TopK}_N(s_{\mathcal G}), \qquad \forall\, \mathcal G \in \mathrm{Groups}(M),
% \]
% where $\mathrm{TopK}_N(s_{\mathcal G})$ returns a binary vector that sets the $N$ largest-score entries in group $\mathcal G$ to 1 and the remaining $M-N$ entries to 0. This target is not used as the forward mask directly; instead, it serves as a curvature-guided direction for updating the soft mask while already respecting the final group-wise structure.
%
% Directly overwriting the soft mask with $t$ would destroy mask continuity and reintroduce premature hard-projection instability. We therefore use $t$ only as a smooth update direction. Let $\delta = m - t$. We update a gate variable $G$ by
% \[
% G \leftarrow G - \eta_{\mathrm{pen}}\,\delta,
% \]
% and then update the mask with an exponential moving average:
% \[
% m \leftarrow (1-\alpha) \cdot m + \alpha \cdot \mathrm{clamp}(G,0,1).
% \]
% This rule preserves soft-mask continuity while moving the mask toward entries that are important under the Hessian-aware loss signal.
% --- REVISED ---
To respect the final semi-structured constraint while keeping the mask continuous, we turn $s$ into a soft group-wise gate rather than directly projecting it onto the feasible set. For an $N$:$M$ pattern, let $\mathrm{Groups}(M)$ denote the collection of non-overlapping groups, each containing $M$ mask entries along the structured sparsity dimension. For a group $\mathcal G$, $s_{\mathcal G}$ denotes the restriction of the score matrix $s$ to the entries in $\mathcal G$. We first standardize the scores layer-wise so that the soft threshold has a stable physical scale, $\tilde s \leftarrow (s - \mu_s)/(\sigma_s + \epsilon)$, and then define, within each group, a soft gate and its hard top-$N$ counterpart:
\begin{equation}
G_{\mathcal G} \;=\; \sigma\!\left(\frac{\tilde s_{\mathcal G} - \tau_{\mathcal G}}{T}\right),
\qquad
\bar G_{\mathcal G} \;=\; \mathrm{TopK}_N(\tilde s_{\mathcal G}),
\qquad \forall\, \mathcal G \in \mathrm{Groups}(M),
\label{eq:soft_gate}
\end{equation}
where $\tau_{\mathcal G}$ is the $N$-th largest entry of $\tilde s_{\mathcal G}$, $\sigma$ is the sigmoid, and $T>0$ is a temperature. The soft gate $G$ already encodes the N:M competition: the $N$ largest entries in each group sit above $0.5$ and the remaining $M-N$ sit below, with the decisiveness controlled by $T$. The hard counterpart $\bar G \in \{0,1\}^{D\times C}$ is kept on the side as a curvature-guided structured target that we will mix into $G$ progressively (\S\ref{sec:progress}).

Directly overwriting the soft mask with $\bar G$ would destroy mask continuity and reintroduce premature hard-projection instability. We therefore use $\bar G$ only as an update direction, and push $G$ toward it through the mid-penalty residual $\delta = m - \bar G$:
\begin{equation}
\tilde G \;=\; \mathrm{clamp}\!\left(G - \eta_{\mathrm{pen}}\,\lambda_{\mathrm{mid}}\,\delta,\; 0,\; 1\right),
\label{eq:mid_inject}
\end{equation}
so that entries that the structured target wants to keep ($\bar G_{ij}=1$) but the current mask has dropped ($m_{ij}$ small) are nudged up, and vice versa. This is precisely the gradient of the binary-preference regularizer $\mathcal L_{\mathrm{mid}}$ (\S\ref{sec:progress}) evaluated against the structured target, and it is the only channel through which $\mathcal L_{\mathrm{mid}}$ influences the mask. Finally, we update the soft mask with an exponential moving average:
\begin{equation}
m \;\leftarrow\; (1-\alpha)\cdot m \;+\; \alpha\cdot \tilde G.
\label{eq:mask_ema}
\end{equation}
This rule preserves soft-mask continuity while moving the mask toward entries that are important under the Hessian-aware loss signal and already compatible with the final N:M structure.

\subsection{Progressive Soft-to-Hard Mask Quenching}
\label{sec:progress}

(I2) implies that an abrupt projection from soft to binary mask is harmful. We therefore introduce \emph{progressive quenching}, a strategy that incorporates structure-aware regularization and penalty terms throughout the heating stage, encouraging the mask to progressively approach a near-binary configuration before the final hardening step.

% Instead, we progressively temper it toward binarization throughout retraining. This progressive process reduces the soft-to-hard gap: early in training, the mask remains continuous and flexible; later, the selected entries are gradually pushed toward 1 and the pruned entries toward 0, so the final projection becomes a mild quenching step rather than an abrupt decision.

First, we introduce a binary-preference regularizer that penalizes ambiguous mask values:
\[
\mathcal L_{\mathrm{mid}} = \frac{1}{|m|}\sum_{i,j} m_{ij}(1-m_{ij}).
\]
This term is maximized at $m_{ij}=0.5$ and minimized at $m_{ij}\in\{0,1\}$. We anneal its weight $\lambda_{\mathrm{mid}}$ during training, keeping the binary preference weak during early exploration and progressively strengthening it as the mask approaches the final deployment stage.

% --- ORIGINAL ---
% The tempering process must also remain structure-aware. Therefore, we progressively blend the current gate with the Hessian-guided structured target:
% \[
% \bar G_{\mathcal G} = \mathrm{TopK}_N(s_{\mathcal G}), \qquad
% G_{\mathrm{final}} = (1-\beta)G + \beta\bar G.
% \]
% Here $\beta$ is annealed from $0$ to $1$, so the update gradually shifts from soft exploration to structured hardening. Consequently, the mask approaches a near-binary state while also aligning with the correct top-$N$ pattern within each group.
% --- REVISED ---
Second, to inject structure awareness during training, we progressively blend the soft gate $G$ from \eqref{eq:soft_gate} with its Hessian-guided structured target $\bar G$ before the mid-penalty injection of \eqref{eq:mid_inject}:
\[
G \;\leftarrow\; (1-\beta)\,G \;+\; \beta\,\bar G,
\qquad \beta: 0 \to 1.
\]
We schedule $\beta$ with a smooth-step (cubic Hermite) curve so that the update gradually shifts from soft exploration to structured hardening without a discontinuity in the gradient of the schedule itself. In parallel, we geometrically anneal the sigmoid temperature in \eqref{eq:soft_gate}, $T \leftarrow \gamma\,T$ with $\gamma\!\in\!(0,1)$ applied at every mask-update step, so that the soft threshold sharpens monotonically as retraining proceeds. Among the three handles, the interpolation controlled by $\beta$ is the primary driver: it continuously morphs the gate from the soft, exploratory $G$ into the hard, group-structured target $\bar G$, while $T$ sharpens the decisiveness of $G$ in sync with this interpolation, and $\lambda_{\mathrm{mid}}$ adds a mild element-wise pressure that pushes any residual mid-range values in $m$ toward $\{0,1\}$. Consequently, the mask approaches a near-binary state while also aligning with the correct top-$N$ pattern within each group. 

Once the soft mask is sufficiently separated, the final quenching step interpolates between the soft mask and its hard version:
\[
m_{\mathrm{eff}} = x\,m + (1-x)\,\mathbf 1[m > \theta],
\]
% --- ORIGINAL ---
% where $x$ is annealed from $1$ to $0$. We then apply the final structured projection, freeze the binary mask, and run a short finetuning stage to absorb the remaining projection error. The overall SparseForge procedure is summarized in Algorithm~\ref{algo:training}.
% --- REVISED ---
where $x$ is annealed linearly from $1$ to $0$ over a dedicated hardening window. Once $x$ reaches $0$, we apply the final structured projection, freeze the binary mask, and run a short finetuning stage to absorb the remaining projection error.

%% file: sections/experiments.tex
\section{Experiments}
\label{sec:exp}

\subsection{Experimental Setup}

\paragraph{Models.}
We evaluate SparseForge on a diverse set of pretrained language models spanning
different architectures and scales, including GPT-2~\citep{radford2019gpt2},
OPT~\citep{zhang2022opt}, LLaMA-2~\citep{touvron2023llama2}, Qwen3~\citep{yang2025qwen3},
and DeepSeek-MoE~\citep{dai2024deepseekmoe}. The evaluated model sizes range from 124M to 16B parameters,
covering both dense transformer models and mixture-of-experts (MoE) architectures.

\paragraph{Hardware setting.}
All experiments are conducted on a cluster of 32 NVIDIA L20A GPUs.
Training is implemented with distributed data parallelism.
For LLaMA-2-7B, the full SparseForge training with 5B tokens requires approximately 50 GPU hours.
Unless otherwise stated, all experiments target 2:4 semi-structured sparsity,
i.e., 50\% sparsity within every group of four weights.
This sparsity pattern is directly supported by modern sparse Tensor Core hardware~\citep{nvidia2020ampere},
making it a practically relevant deployment setting.

\paragraph{Training setup.}
For LLaMA-2-7B, we perform sparse retraining on Dolmino-mix-1124~\citep{olmo2,dolmino1124},
following the data configuration adopted in recent continual pretraining and sparse
recovery studies. For all other models, we use the C4 corpus~\citep{raffel2020t5,dodge2021c4}.
Unless otherwise specified, we use a total retraining budget of approximately
5B tokens for LLaMA-2-7B and 1.25B tokens for other models, substantially
smaller than prior retraining-based methods such as AST~\citep{huang2025ast} and CAST~\citep{huang2025cast}. Unless otherwise specified, each SparseForge configuration is run with three random seeds. We report the mean performance across the three runs. 

\begin{table*}[b!]
\centering
\caption{Compact cross-model summary under 2:4 semi-structured sparsity. We report mean zero-shot accuracy (\%) for the dense model, SparseForge, and their difference. Detailed task-level results are provided in Appendix Table~\ref{tab:main_results_full}.}
\label{tab:main_results_summary}
\resizebox{\textwidth}{!}{%
\begin{tabular}{l|cccccccc}
\toprule
\textbf{Metric} & \textbf{GPT2-M} & \textbf{GPT2-L} & \textbf{GPT2-XL} & \textbf{OPT-2.7B} & \textbf{Qwen3-1.7B} & \textbf{Qwen3-8B} & \textbf{Qwen3-14B} & \textbf{DeepSeek-MoE} \\
\midrule
Dense       & 40.97 & 42.76 & 45.49 & 47.76 & 56.51 & 65.73 & 68.36 & 59.54 \\
SparseForge & 40.31 & 42.10 & 44.34 & 46.67 & 53.33 & 63.31 & 65.44 & 58.57 \\
Diff        & -0.66 & -0.66 & -1.15 & -1.09 & -3.18 & -2.42 & -2.93 & -0.97 \\
\bottomrule
\end{tabular}%
}
\end{table*}

\paragraph{Evaluation metrics.}
We report WikiText-2 perplexity (PPL) and zero-shot accuracy on downstream tasks
evaluated with \texttt{lm-evaluation-harness}~\citep{biderman2024lmeval}.
For the cross-model comparison in Table~\ref{tab:main_results_summary}, we use a benchmark
suite that is consistent across model families and remains aligned with the reporting
protocol of recent semi-structured sparsification work. This suite includes
ARC-Challenge and ARC-Easy~\citep{clark2018arc},
BoolQ~\citep{clark2019boolq},
HellaSwag~\citep{zellers2019hellaswag},
OpenBookQA~\citep{mihaylov2018openbookqa},
RTE~\citep{wang2019superglue},
and WinoGrande~\citep{sakaguchi2021winogrande}.

For the detailed LLaMA-2-7B comparison in Table~\ref{tab:llama2_compare},
we follow the benchmark configuration of CAST~\citep{huang2025cast}, adding
RACE~\citep{lai2017race} and PIQA~\citep{bisk2020piqa} to the above suite.
We report mean accuracy over the corresponding task set in each table.

\subsection{Results Across Model Families}

To facilitate consistent comparison across architectures while staying aligned
with recent semi-structured sparsification practice, we summarize the cross-model
results in Table~\ref{tab:main_results_summary}, while detailed task-level results
are deferred to Appendix Table~\ref{tab:main_results_full}. SparseForge preserves
model quality across diverse architectures and scales: although degradation varies
across families, the drops in mean accuracy remain moderate in most cases,
showing that continuous mask optimization with progressive quenching is a robust
recipe for semi-structured sparsification beyond a single model family.

\begin{table*}[t!]
\centering
\caption{Comparison with prior semi-structured sparsification methods on LLaMA-2-7B under 2:4 sparsity. Best results among \emph{sparse methods} are highlighted in \textbf{bold}. For dense and SparseForge rows, we report our own re-evaluation results under the current
\texttt{lm-evaluation-harness} setup; prior baseline numbers are taken from the
corresponding original papers unless otherwise noted. $^\dagger$ denotes a continual-training variant using a substantially larger retraining token budget.}
\label{tab:llama2_compare}
\resizebox{\textwidth}{!}{%
\begin{tabular}{lcc|ccccccc|c|c}
\toprule
\textbf{Method} & \textbf{Pattern} & \textbf{Tokens} & \textbf{HellaS.} & \textbf{RACE} & \textbf{PIQA} & \textbf{WinoG.} & \textbf{ARC-e} & \textbf{ARC-c} & \textbf{OBQA} & \textbf{Average} & \textbf{Wiki PPL} \\
\midrule
Dense         & Dense & 2T   & 57.18 & 39.52 & 78.07 & 69.06 & 76.26 & 43.52 & 31.40 & 56.43 & 5.12 \\
CAST$^\dagger$ & 2:4  & 40B  & 56.13 & 40.86 & 77.58 & 69.53 & 77.78 & 47.18 & 33.60 & 57.52 & 5.21 \\
\midrule
Wanda         & 2:4   & $\times$ & 41.05 & 35.02 & 70.78 & 62.67 & 61.99 & 27.56 & 22.80 & 45.98 & 11.29 \\
SparseGPT     & 2:4   & $\times$ & 43.36 & 36.84 & 71.38 & 63.69 & 62.84 & 29.18 & 22.80 & 47.16 & 10.42 \\
MaskLLM       & 2:4   & 2B   & 50.91 & 40.77 & 74.92 & 64.48 & 69.57 & 36.00 & 28.00 & 52.09 & 6.72 \\
Naive Retrain & 2:4   & 10B  & 53.90 & 38.28 & 76.61 & 68.27 & 75.21 & 41.21 & 29.60 & 54.73 & 5.78 \\
SR-STE        & 2:4   & 10B  & 54.02 & 39.02 & 76.88 & 68.35 & 75.58 & 41.46 & 29.60 & 54.99 & 5.74 \\
AST           & 2:4   & 7.5B & 54.66 & -- & -- & 67.87 & 73.61 & 39.93 & 28.60 & -- & -- \\
AST + SLoRB   & 2:4   & 7.5B & \textbf{55.24} & -- & -- & 68.48 & 74.91 & 41.11 & 29.40 & -- & -- \\
CAST          & 2:4   & 7.5B & 54.50 & 40.48 & \textbf{77.09} & 68.27 & 76.52 & 43.68 & 30.80 & 55.91 & \textbf{5.58} \\
SparseForge (ours) & 2:4 & 1.25B & 52.95 & 41.15 & 75.46 & 69.14 & 76.35 & 43.86 & 32.80 & 55.96 & 6.24 \\
SparseForge$^\dagger$ (ours) & 2:4   & 5B   & 54.42 & \textbf{41.63} & 76.88 & \textbf{69.69} & \textbf{77.95} & \textbf{45.31} & \textbf{35.00} & \textbf{57.27} & 6.09 \\
\bottomrule
\end{tabular}%
}
\end{table*}

\subsection{Comparison with Prior Methods on LLaMA-2-7B}

We compare SparseForge with representative post-training and retraining-based
semi-structured sparsification methods, including SparseGPT~\citep{frantar2023sparsegpt},
Wanda~\citep{sun2023wanda}, MaskLLM~\citep{fang2024maskllm},
AST~\citep{huang2025ast}, and CAST~\citep{huang2025cast}, under aligned 2:4 sparsity.

\begin{figure}[t!]
\centering
\includegraphics[width=\linewidth]{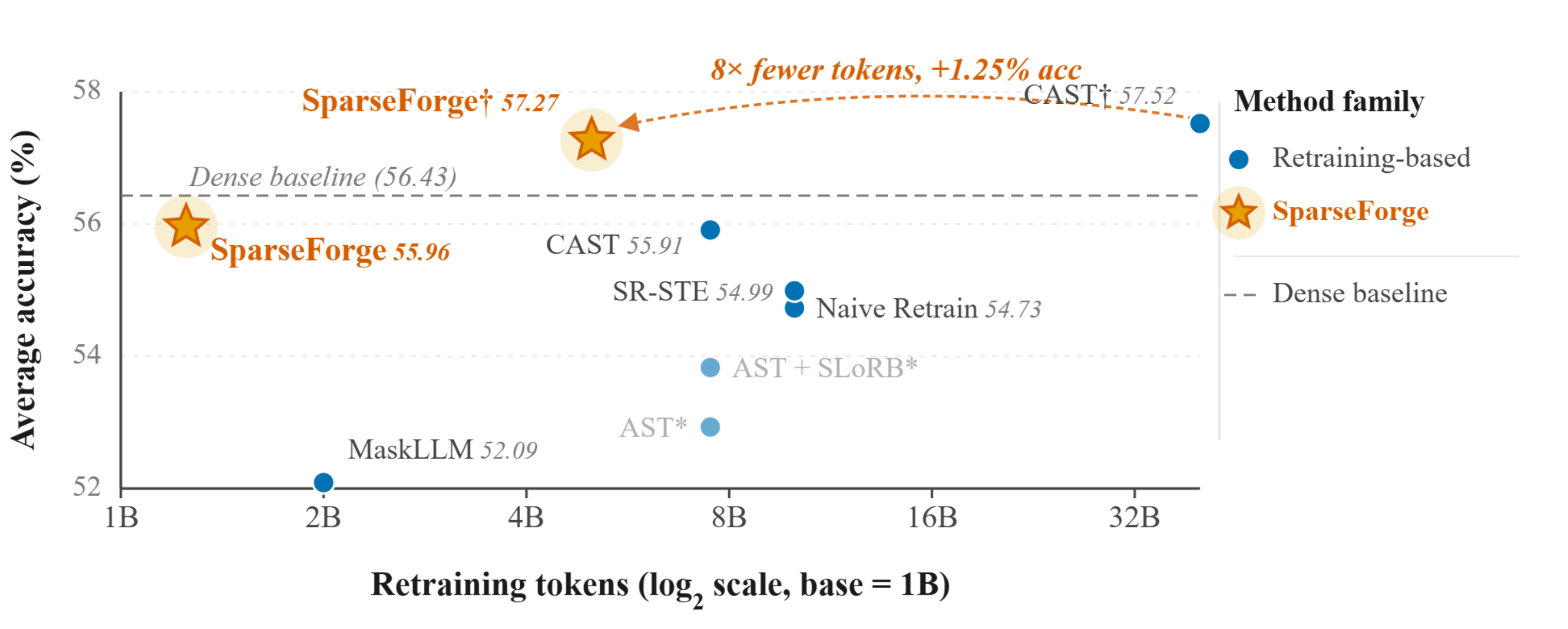}
\caption{Accuracy vs.\ retraining tokens on LLaMA-2-7B (2:4 sparsity, log-scale).
SparseForge matches the 40B-token CAST$^\dagger$ with $\sim$8$\times$ fewer tokens.
AST/AST+SLoRB (semi-transparent) report only a 5-task mean.}
\label{fig:tokens_vs_accuracy}
\end{figure}

As shown in Table~\ref{tab:llama2_compare} and Figure~\ref{fig:tokens_vs_accuracy},
SparseForge achieves 57.27\% average
accuracy using only 5B retraining tokens, outperforming the 7.5B-token CAST
result (55.91\%) and approaching the stronger 40B-token continual-training CAST
variant (57.52\%) with roughly $8\times$ fewer retraining tokens.
This indicates that improved mask optimization and structure-aware
tempering can substantially reduce the retraining required for high-quality
semi-structured sparse recovery.

\begin{table*}[h]
\centering
\caption{Full ablation on LLaMA-2-7B under 2:4 sparsity. Unless otherwise noted, variants are trained on C4 with 1.25B retraining tokens (20k steps). We report WikiText-2 perplexity, mean zero-shot accuracy, and the difference against the full SparseForge configuration.}
\label{tab:ablation_full}
\resizebox{\textwidth}{!}{%
\begin{tabular}{lcccc}
\toprule
\textbf{Variant} & \textbf{Steps / Tokens} & \textbf{Wiki PPL} & \textbf{Mean Acc.} & \textbf{Diff vs Full} \\
\midrule
Dense                                & 2T           & 5.1983 & 59.71\% & +2.48\% \\
SparseForge (full)                   & 20k / 1.25B  & 6.3823 & 57.23\% & -- \\
\midrule
SparseForge (dolmino)                & 20k / 1.25B     & 6.1037 & 59.22\% & +1.99\% \\
SparseForge w/ 10k steps             & 10k / 0.625B & 6.4755 & 55.89\% & -1.34\% \\
SparseForge w/ magnitude             & 20k / 1.25B  & 6.5275 & 55.61\% & -1.62\% \\
SparseForge w/o SLoRB                & 20k / 1.25B  & 6.3617 & 55.04\% & -2.19\% \\
SparseForge w/ larger teacher (13B)  & 20k / 1.25B  & 6.6089 & 54.87\% & -2.36\% \\
SparseForge w/o distillation         & 20k / 1.25B  & 6.6594 & 55.72\% & -1.51\% \\
\bottomrule
\end{tabular}%
}
\end{table*}

\subsection{Ablation Studies}
\label{sec:exp:ablation}

We conduct a full ablation of SparseForge on LLaMA-2-7B under 2:4 sparsity, using C4~\citep{raffel2020t5,dodge2021c4} as the default retraining corpus with 1.25B retraining tokens (20k steps). Table~\ref{tab:ablation_full} studies the effects of retraining data, training budget, importance criterion, SLoRB, teacher scale, and distillation. 

Several trends are clear. First, retraining data quality matters substantially: replacing C4 with Dolmino-mix-1124 improves mean accuracy from 57.23\% to 59.22\% and also lowers WikiText-2 perplexity. Second, SparseForge remains relatively token-efficient: halving the budget from 20k to 10k steps reduces mean accuracy by only 1.34\%. Third, Hessian-aware importance is important for resolving intra-group competition: replacing it with magnitude-based scoring lowers mean accuracy by 1.62\% and worsens perplexity. Finally, the remaining components also make nontrivial contributions: removing SLoRB causes the largest drop among these auxiliary ablations (-2.19\%), while removing distillation (-1.51\%) or replacing the teacher with a larger 13B model (-2.36\%) also degrades recovery. Overall, these results support the view that SparseForge benefits from both better mask optimization and careful training design, rather than from retraining scale alone.
% \begin{table}[htbp]
% \centering
% \caption{Compact ablation on LLaMA-2-7B under 2:4 sparsity. We vary one axis at a time against the full SparseForge configuration (20k steps, 1.25B tokens on C4, Hessian-aware importance) and report WikiText-2 perplexity and mean zero-shot accuracy. The full ablation is provided in Appendix~Table~\ref{tab:ablation_full}.}
% \label{tab:ablation}
% \small
% \begin{tabular}{lcccc}
% \toprule
% \textbf{Variant} & \textbf{Steps / Tokens} & \textbf{Wiki PPL} & \textbf{Mean Acc.} & \textbf{Diff vs Full} \\
% \midrule
% Dense                           & 2T           & 5.1983 & 59.71\% & +2.48\% \\
% SparseForge (full)              & 20k / 1.25B  & 6.3823 & 57.23\% & -- \\
% \midrule
% SparseForge (dolmino)           & 20k / 1.25B     & 6.1037 & 59.22\% & +1.99\% \\
% SparseForge w/ 10k steps        & 10k / 0.625B & 6.4755 & 55.89\% & -1.34\% \\
% SparseForge w/ magnitude        & 20k / 1.25B  & 6.5275 & 55.61\% & -1.62\% \\
% \bottomrule
% \end{tabular}
% \end{table}
% 如果地方不够的话可以只留下dolmino, 10k steps, maginutde.剩下送到appendix里面

%% file: sections/conclusions.tex
\section{Conclusion}

We introduced \textbf{SparseForge}, a post-training framework for semi-structured sparsification that follows a mask-annealing pipeline: SparseForge explicitly learns and progressively quenches the sparsity mask toward a deployable 2:4 pattern, driven by a dual-track retraining loop that jointly updates weights and the soft mask under a Hessian-guided structured target. Our results suggest that under grouped semi-structured constraints, sparse recovery is governed more by mask quality than by retraining scale alone. By combining explicit soft-mask optimization, Hessian-guided structured targets, and progressive quenching within this dual-track loop, SparseForge achieves strong sparse recovery across model families and a favorable accuracy--efficiency trade-off under semi-structured sparsity, matching or surpassing prior state-of-the-art recovery while using substantially fewer retraining tokens. More broadly, these findings highlight mask-centric optimization as a promising direction for hardware-aligned sparse LLM compression.

%% file: sections/appendix.tex
\newpage
\appendix

\section{Detailed Cross-Model Results Under 2:4 Sparsity}

% \subsection{Detailed Cross-Model Results Under 2:4 Sparsity}

For readability, the main text reports a compact cross-model summary using mean
zero-shot accuracy only. In this appendix, we provide the full task-level results
corresponding to the cross-model comparison in Table~\ref{tab:main_results_summary}.
Table~\ref{tab:main_results_full} reports the dense-model performance, SparseForge
performance, and their differences on each benchmark under 2:4 semi-structured
sparsity, together with the mean accuracy and WikiText-2 perplexity.
\begin{table*}[htbp]
\centering
\caption{Detailed cross-model results under 2:4 semi-structured sparsity. We report zero-shot accuracy (\%) on downstream tasks, the mean accuracy across tasks, and WikiText-2 perplexity (PPL) for the dense model, SparseForge, and their differences.}
\label{tab:main_results_full}
\resizebox{\textwidth}{!}{%
\begin{tabular}{ll|ccccccc|c|c}
\toprule
\textbf{Model} & \textbf{Variant} & \textbf{ARC-c} & \textbf{ARC-e} & \textbf{BoolQ} & \textbf{Hella.} & \textbf{OBQA} & \textbf{RTE} & \textbf{Wino.} & \textbf{Mean} & \textbf{Wiki PPL} \\
\midrule

\multirow{3}{*}{GPT-2-Medium}
& Dense  & 21.67 & 49.07 & 58.41 & 33.23 & 18.60 & 52.71 & 53.12 & 40.97 & 21.80 \\
& SparseForge & 21.16 & 46.04 & 60.09 & 31.70 & 16.40 & 53.43 & 53.35 & 40.31 & 25.90 \\
& Diff   & -0.51 & -3.03 & +1.68 & -1.53 & -2.20 & +0.72 & +0.24 & -0.66 & +4.10 \\

\midrule

\multirow{3}{*}{GPT-2-Large}
& Dense  & 21.84 & 53.16 & 60.55 & 36.39 & 19.40 & 52.71 & 55.25 & 42.76 & 18.95 \\
& SparseForge & 22.10 & 50.29 & 61.90 & 34.68 & 19.40 & 52.71 & 53.59 & 42.10 & 23.57 \\
& Diff   & +0.26 & -2.86 & +1.35 & -1.70 & +0.00 & +0.00 & -1.66 & -0.66 & +4.62 \\

\midrule

\multirow{3}{*}{GPT-2-XL}
& Dense  & 25.09 & 58.29 & 61.68 & 40.00 & 22.60 & 52.35 & 58.41 & 45.49 & 17.12 \\
& SparseForge & 23.29 & 55.98 & 61.77 & 38.37 & 21.40 & 54.15 & 55.41 & 44.34 & 20.53 \\
& Diff   & -1.79 & -2.31 & +0.09 & -1.63 & -1.20 & +1.81 & -3.00 & -1.15 & +3.41 \\

\midrule

\multirow{3}{*}{OPT-2.7B}
& Dense  & 26.79 & 60.77 & 60.40 & 45.86 & 25.00 & 54.51 & 61.01 & 47.76 & 12.45 \\
& SparseForge & 25.17 & 58.67 & 63.06 & 44.44 & 21.80 & 52.71 & 60.85 & 46.67 & 15.14 \\
& Diff   & -1.62 & -2.10 & +2.66 & -1.41 & -3.20 & -1.81 & -0.16 & -1.09 & +2.69 \\

\midrule

\multirow{3}{*}{Qwen3-1.7B}
& Dense  & 39.85 & 72.26 & 77.58 & 46.09 & 28.20 & 70.76 & 60.85 & 56.51 & 15.56 \\
& SparseForge & 37.97 & 70.08 & 74.53 & 46.30 & 27.00 & 56.68 & 60.77 & 53.33 & 10.54 \\
& Diff   & -1.88 & -2.19 & -3.06 & +0.21 & -1.20 & -14.08 & -0.08 & -3.18 & -5.02 \\

\midrule

\multirow{3}{*}{Qwen3-8B}
& Dense  & 55.80 & 83.50 & 86.61 & 57.14 & 31.00 & 78.34 & 67.72 & 65.73 & 8.97 \\
& SparseForge & 50.85 & 81.52 & 82.35 & 54.34 & 31.60 & 72.92 & 69.30 & 63.31 & 8.03 \\
& Diff   & -4.95 & -1.98 & -4.25 & -2.80 & +0.60 & -5.42 & +1.58 & -2.42 & -0.94 \\

\midrule

\multirow{3}{*}{Qwen3-14B}
& Dense  & 58.45 & 84.18 & 89.33 & 60.96 & 35.00 & 77.62 & 73.01 & 68.36 & 7.89 \\
& SparseForge & 53.67 & 82.24 & 84.92 & 57.94 & 34.40 & 72.20 & 72.69 & 65.44 & 7.17 \\
& Diff   & -4.78 & -1.94 & -4.40 & -3.03 & -0.60 & -5.42 & -0.32 & -2.93 & -0.72 \\

\midrule

\multirow{3}{*}{DeepSeek-MoE}
& Dense  & 45.05 & 75.88 & 72.72 & 58.09 & 32.00 & 62.82 & 70.24 & 59.54 & 6.28 \\
& SparseForge & 41.47 & 74.12 & 73.82 & 57.45 & 32.00 & 62.09 & 69.06 & 58.57 & 6.92 \\
& Diff   & -3.58 & -1.77 & +1.10 & -0.64 & +0.00 & -0.72 & -1.18 & -0.97 & +0.64 \\

\bottomrule
\end{tabular}%
}
\end{table*}
\section{Extension to Block-16 Sparsity}
\label{app:block16}

Although our main experiments focus on the practically important 2:4 semi-structured
pattern, the SparseForge formulation is not inherently restricted to this setting. In
particular, the score-based projection and progressive quenching mechanism can be
adapted to other structured sparsity patterns by redefining the feasible group and
projection operator. To examine this generality, we additionally evaluate SparseForge
under a block-16 sparsity pattern. Concretely, in our block-16 setting, each
$16\times16$ weight block is constrained to contain only 50\% nonzero entries.
Compared with 2:4 sparsity, this is a relatively looser form of structured sparsity:
the sparsity pattern is still constrained at the block level, but the competition
among weights is imposed over a larger local region rather than by a strict fine-grained
$n\!:\!m$ rule within every group of four.

Table~\ref{tab:block16_results} reports results on several representative model
families under block-16 sparsity. Overall, SparseForge remains effective beyond 2:4,
with moderate degradation on most model families and benchmarks. In particular,
LLaMA-2-7B retains strong performance under block-16, with mean accuracy dropping
from 59.70\% to 59.02\% (a 0.68-point decrease). OPT-2.7B remains nearly unchanged,
showing a slight mean improvement from 47.76\% to 47.93\%. Similar trends are also
observed on GPT-2-XL, where the mean accuracy only decreases from 45.49\% to
45.14\%.

At the same time, the block-16 setting appears to be more challenging for some
architectures, especially smaller or more sensitive models. For example, Qwen3-1.7B
shows a larger drop in mean accuracy (from 56.47\% to 53.49\%), and DeepSeek-MoE-16B
also exhibits a noticeable decrease (from 59.54\% to 58.20\%). These results suggest
that while the proposed mask optimization strategy transfers beyond 2:4, the degree
of recoverability under block-structured sparsity still depends on model architecture
and training dynamics.

Overall, these results support the view that SparseForge is a general mask optimization
framework rather than a method tied specifically to 2:4 sparsity. We leave a more
systematic study of different structured sparsity patterns and their hardware--accuracy
\label{app:block16}
\begin{table}[htbp]
\centering
\caption{SparseForge under block-16 sparsity across model families. We report mean zero-shot accuracy before and after sparsification.}
\label{tab:block16_results}
\small
\begin{tabular}{lccc}
\toprule
\textbf{Model} & \textbf{Base Mean} & \textbf{Block-16 Mean} & \textbf{Diff} \\
\midrule
LLaMA-2-7B        & 59.70 & 59.02 & -0.68 \\
OPT-2.7B          & 47.76 & 47.93 & +0.17 \\
GPT-2             & 37.13 & 37.58 & +0.45 \\
GPT-2-Medium      & 41.01 & 39.45 & -1.56 \\
GPT-2-Large       & 42.65 & 41.87 & -0.81 \\

GPT-2-XL          & 45.49 & 45.14 & -0.35 \\
Qwen3-1.7B        & 56.47 & 53.49 & -2.98 \\
DeepSeek-MoE-16B  & 59.54 & 58.20 & -1.34 \\
\bottomrule
\end{tabular}
\end{table}
\begin{table*}[htbp]
\centering
\caption{Detailed results of SparseForge under block-16 sparsity on LLaMA-2-7B.}
\label{tab:block16_llama2}
\resizebox{\textwidth}{!}{%
\begin{tabular}{l|ccccccc|c}
\toprule
\textbf{Variant} & \textbf{ARC-c} & \textbf{ARC-e} & \textbf{BoolQ} & \textbf{Hella.} & \textbf{OBQA} & \textbf{RTE} & \textbf{Wino.} & \textbf{Mean} \\
\midrule
Dense    & 43.43 & 76.26 & 77.71 & 57.17 & 31.40 & 62.82 & 69.14 & 59.70 \\
SparseForge   & 41.89 & 75.25 & 76.24 & 56.78 & 32.20 & 61.73 & 69.06 & 59.02 \\
Diff     & -1.54 & -1.01 & -1.47 & -0.39 & +0.80 & -1.08 & -0.08 & -0.68 \\
\bottomrule
\end{tabular}%
}
\end{table*}
% \begin{figure}[htbp]
%     \centering
%     \makebox[\textwidth][c]{
%     \includegraphics[width=1\linewidth]{images/SparseForge_motivation.png}
% } 
%     \caption{Figure. Motivation for SparseForge. (a) Conventional methods compute importance scores from a single Hessian snapshot and immediately commit to a hard binary mask. This one-shot estimate is inherently approximate; once locked in, errors in the mask cannot be corrected, leading to suboptimal final performance. (b) SparseForge trains a continuous soft mask alongside the model. Because the mask is differentiable, gradients accumulate over many training steps, allowing importance estimates to self-correct — weights initially deemed unimportant may be restored, and vice versa. The mask converges to the true optimal 2:4 structure before final hard projection. (Bottom) Both methods see a PPL rise after pruning begins, followed by gradual recovery during finetuning. However, hard pruning (red) plateaus at a higher PPL due to its inaccurate one-shot mask, while SparseForge (green) converges to a lower PPL thanks to more precise importance captured by accumulated gradients.}
%     \label{fig:motivation}
% \end{figure} 
\section{End-to-End Inference Speedup}
\label{app:speedup}
\begin{table}[htbp]
\centering
\small
\caption{End-to-end inference throughput (tokens/s) of SparseForge 2:4 sparse LLaMA-2-7B vs.\ dense baseline, measured with TensorRT-LLM on an NVIDIA H800 GPU under different input/output sequence lengths. The last row reports the model weight memory footprint. ``Efficiency'' denotes the sparse-over-dense ratio (speedup for throughput, compression ratio for memory).}
\label{tab:speedup_trtllm}
\begin{tabular}{l|ccc}
\toprule
\textbf{(Inp Len, Out Len)} & \textbf{Sparse} & \textbf{Dense} & \textbf{Efficiency} \\
\midrule
(128, 128)    & 108.72 & 75.41 & 1.44$\times$ \\
(128, 1024)   & 106.18 & 74.63 & 1.42$\times$ \\
(1024, 128)   & 103.87 & 73.62 & 1.41$\times$ \\
(1024, 1024)  & 101.29 & 72.38 & 1.40$\times$ \\
\midrule
\textbf{Memory Consumption} & 7.31\,GB & 12.55\,GB & 0.58$\times$ \\
\bottomrule
\end{tabular}
\end{table}
Beyond accuracy recovery, a practical motivation for adopting the 2:4
semi-structured pattern is that it is directly supported by modern NVIDIA sparse
Tensor Cores~\citep{nvidia2020ampere}, enabling tangible end-to-end acceleration
on real hardware. To verify that our SparseForge-trained 2:4 sparse models
translate into actual deployment gains, we evaluate their inference throughput
using TensorRT-LLM\footnote{\url{https://github.com/NVIDIA/TensorRT-LLM}} on an
NVIDIA H800 GPU, under four different input/output sequence length settings.
We use the 2:4 sparse LLaMA-2-7B model produced by SparseForge, and compare its
decoding throughput (tokens/s) and memory footprint against the corresponding
dense model under identical inference configurations.

Table~\ref{tab:speedup_trtllm} reports the results. Across all input/output
length configurations, the 2:4 sparse model consistently outperforms its dense
counterpart, delivering $1.40\times$--$1.44\times$ throughput improvements on
H800. In addition, the memory footprint of model weights is reduced to roughly
58\% of the dense baseline (7.31\,GB vs.\ 12.55\,GB for LLaMA-2-7B), which
directly lowers the GPU memory requirement for deployment.
Together with the accuracy results in the main text, these measurements confirm
that SparseForge produces 2:4 sparse models that are not only competitive in
quality but also practically deployable, converting the algorithmic sparsity
gain into measurable end-to-end latency and memory savings on commodity
inference hardware.

\section{Discussion and Limitations}

\paragraph{Scaling retraining tokens is not the only path.}
Recent semi-structured sparsification work largely treats closing the gap to
the dense model as a matter of \emph{how many tokens} one is willing to spend
on retraining. Our results push back on this implicit assumption: on LLaMA-2-7B
under 2:4 sparsity, SparseForge matches the strongest retraining-based baseline
(CAST$^\dagger$) while using roughly an order of magnitude fewer retraining
tokens (Table~\ref{tab:llama2_compare}, Figure~\ref{fig:tokens_vs_accuracy}),
and this efficiency does not come at the cost of generality---the same recipe
preserves model quality across architectures ranging from GPT-2 variants to
14B dense Qwen3 and a 16B MoE model (Table~\ref{tab:main_results_summary}).
The ablations in \S\ref{sec:exp:ablation} further show that the gain cannot be
attributed to a single trick: removing Hessian-aware importance, the SLoRB
branch, or structure-aware signal each causes a clear drop, indicating that
making the mask a first-class optimization variable, scoring it with
curvature-aware signals, and annealing it into the deployable pattern address
complementary failure modes of semi-structured sparse recovery. This does not
argue against retraining at scale; rather, it suggests that \emph{where}
optimization effort is spent matters at least as much as how long it runs, and
that better mask optimization is a more efficient lever than additional tokens
for pushing semi-structured sparse models toward dense-level quality.

\paragraph{Limitations.}
Our results remain sensitive to the choice of retraining corpus: the larger
degradation observed on Qwen3 in
Table~\ref{tab:main_results_summary} is most plausibly explained by the
mismatch between its reasoning-oriented pretraining mixture and the generic
C4 data we use for recovery, and a stronger task-aligned corpus would likely
narrow this gap. We also focus primarily on 2:4 sparsity---with only a
preliminary block-16 study in Appendix~\ref{app:block16}---on a fixed set of
model families.